\documentclass[sigconf,screen]{acmart}
\usepackage{booktabs} 
\usepackage{algorithm2e}
\usepackage{subfig}
\usepackage{amsmath}
\usepackage{float}
\usepackage{setspace}

\setcopyright{rightsretained}

\acmISBN{978-1-4503-7596-2}

\acmConference[ICVGIP'21]{12th Indian Conference on Computer Vision, Graphics and Image Processing}{December 2021}{Jodhpur, India}
\acmYear{2021}
\copyrightyear{2021}

\acmPrice{15.00}

\editor{Chetan Arora}
\editor{Parag Chaudhuri}
\editor{Subhransu Maji}

\acmDOI{https://doi.org/10.1145/3490035.3490280}

\acmArticle{23}

\begin{document}

\title{Automatic Quantification and Visualization of Street Trees}
\titlenote{Produces the permission block, and
  copyright information}
  \author{Arpit Bahety \hspace{2mm} Rohit Saluja \hspace{2mm} Ravi Kiran Sarvadevabhatla \hspace{2mm} Anbumani Subramanian \hspace{2mm} C.V. Jawahar}
  \email{arpitbahety39@gmail.com, rohit.saluja@research.iiit.ac.in, {ravi.kiran, anbumani, jawahar}@iiit.ac.in}  
      \affiliation{
    \institution{\url{}\\{\bf Centre For Visual Information Technology (CVIT)} \\{\bf International Institute of Information Technology - Hyderabad (IIIT-H)} }
    \city{{\bf Gachibowli, Hyderabad 500032}}
    \country{{\bf INDIA}}
    \postcode{500032}
  }
  
\renewcommand{\shortauthors}{}

\begin{abstract}
Assessing the number of street trees is essential for evaluating urban greenery and can help municipalities employ solutions to identify tree-starved streets. It can also help identify roads with different levels of deforestation and afforestation over time. Yet, there has been little work in the area of street trees quantification. This work first explains a data collection setup carefully designed for counting roadside trees. We then describe a unique annotation procedure aimed at robustly detecting and quantifying trees. We work on a dataset of around $1300$ Indian road scenes annotated with over $2500$ street trees. We additionally use the five held-out videos covering $25$ km of roads for counting trees. We finally propose a street tree detection, counting, and visualization framework using current object detectors and a novel yet simple counting algorithm owing to the thoughtful collection setup. We find that the high-level visualizations based on the density of trees on the routes and Kernel Density Ranking (KDR) provide a quick, accurate, and inexpensive way to recognize tree-starved streets. We obtain a tree detection mAP of $83.74\%$ on the test images, which is a $2.73\%$ improvement over our baseline. We propose Tree Count Density Classification Accuracy (TCDCA) as an evaluation metric to measure tree density. We obtain TCDCA of $96.77\%$ on the test videos, with a remarkable improvement of $22.58\%$ over baseline, and demonstrate that our counting module's performance is close to human level. Source code:~\url{https://github.com/iHubData-Mobility/public-tree-counting}.
\end{abstract}


\maketitle
\begin{figure}[h]
\includegraphics[trim={0cm 0.2cm 0 0},clip,width=\linewidth]{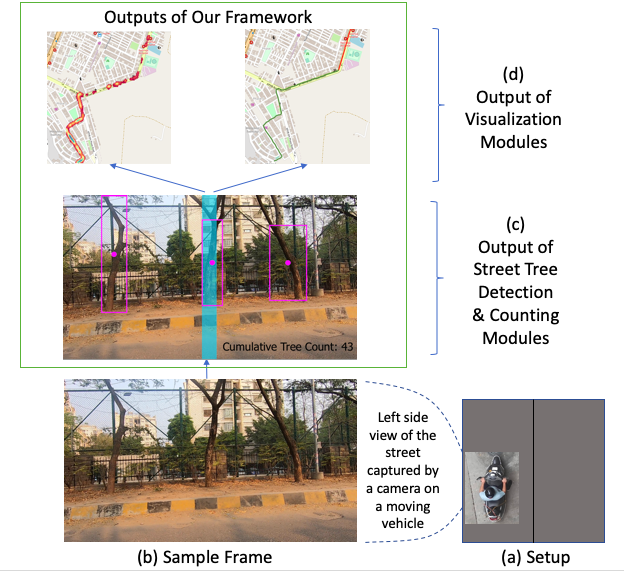}\Description{Moneyshot Figure}
\vspace{-0.5em}
\caption{Output of different modules of our framework for data collection, quantification, and visualization of street trees. Such a framework with city-wide visualizations can help in urban planning. (a) shows data collection platform (explained in Section~\ref{section:setup&dataset}), (b) sample of the captured frame, (c) output from detection and counting modules where detections are shown in pink and counting range by the blue bar, and (d) illustrates the final visualizations based on two different measures of tree density on the routes.}
\label{fig:moneyshot}\vspace{-1em}
\end{figure}
\section{Introduction}
Increased urbanization is destroying natural ecosystems and degrading the environment of urban areas. The services provided by urban trees can mitigate some of these problems~\cite{wolf, roy}. An essential part of urban trees is street trees, which provide numerous services. Studies in tropical cities have found that street trees help reducing air temperature, humidity, and air pollution~\cite{vailshery}. For example, in Bangalore, India, an experimental study showed that afternoon ambient air temperatures were 5.6 °C lower in roads lined with trees, and road surface temperatures 27.5 °C lower than those measured in comparable tree-less streets~\cite{vailshery}. Street trees also reduce flooding and stormwater damage~\cite{vico, stovin, day}, attenuate traffic noise~\cite{kalansuriya}, and increase perceived neighborhood safety~\cite{kostas}. The above indicates that having a system to assess street trees is important. Civic or municipal authorities could use such a system to identify tree-starved streets and employ afforestation efforts. 
Over the years, various works have been developed to quantify urban greenery, detect and catalog trees. However, many of these works rely on aerial or satellite imagery, which is not the fittest setting for assessing street trees~\cite{yangJun}, or they do not provide high-level visualizations, which are imperative for authorities to understand tree coverage and take adequate measures. In this paper, we propose a system to detect and count street trees and showcase the number of street trees in a meaningful way. We also explain our data collection setup and annotation guideline for street trees (refer Section~\ref{section:setup&dataset}). We present two different approaches to visualize the results, which are explained in Sections~\ref{section:categoryMap} and~\ref{section:densityMap}. These visualizations can help achieve a high-level view of the street tree distribution in cities, towns, or villages. The outputs from different modules of the entire framework are briefly explained in Figure~\ref{fig:moneyshot} (refer to Figure~\ref{fig:pipeline} for further details). The core of our system is modules for the detection and counting of street trees. By using our unique street tree annotations and tweaking the YOLOv5 model, we obtain an mAP of $83.74\%$. We then develop a novel counting algorithm to quantify the street trees. We evaluate the street tree counting using a metric defined as tree count density classification accuracy (refer Section~\ref{section:results}) where we obtain an accuracy of $96.77\%$ which is a $22.58\%$ improvement over the baseline.

The key contributions of this work are:
\begin{itemize}
\item A setup carefully designed to count roadside trees and a unique tree annotation strategy, which also helps to solve the intra-class occlusion problem~\cite{xie}. Unlike other works on tree detections, our annotation process is designed to avoid ambiguities in the tree boundaries (refer Section~\ref{section:setup&dataset}). Hence, we achieve highly reliable detections and counting estimates.
\item A framework to quantify street trees based on object detection techniques. We also develop an algorithm to quantify the street trees.
\item High-level visualizations based on the density of trees on the routes and Kernel Density Ranking (KDR).
\end{itemize}
\begin{figure*}
\makebox[\textwidth]{\includegraphics[trim={0cm 0cm 0.2cm 0cm},clip,width=0.9\textwidth]{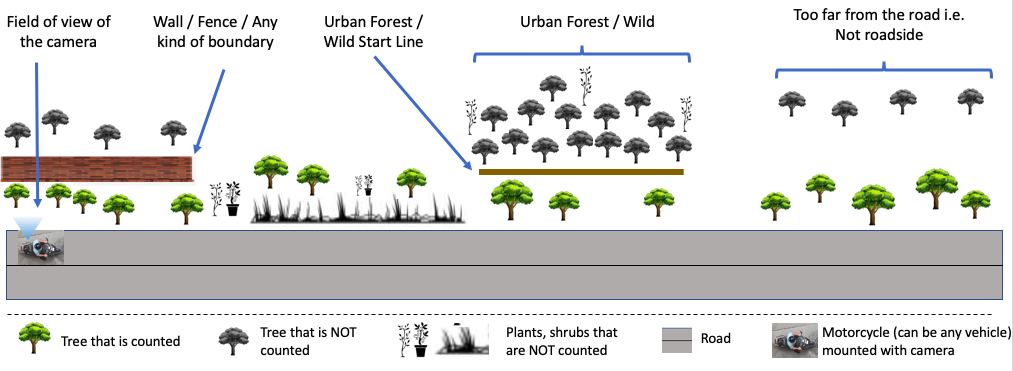}}\Description{Depiction}
\vspace{-1.5em}
\caption{Depiction of trees that are counted and not counted by our system.}
\label{fig:scheme}
\vspace{-1em}
\end{figure*}
\begin{figure*}
    \centering
    \subfloat[\centering]{{\includegraphics[width=0.247\textwidth]{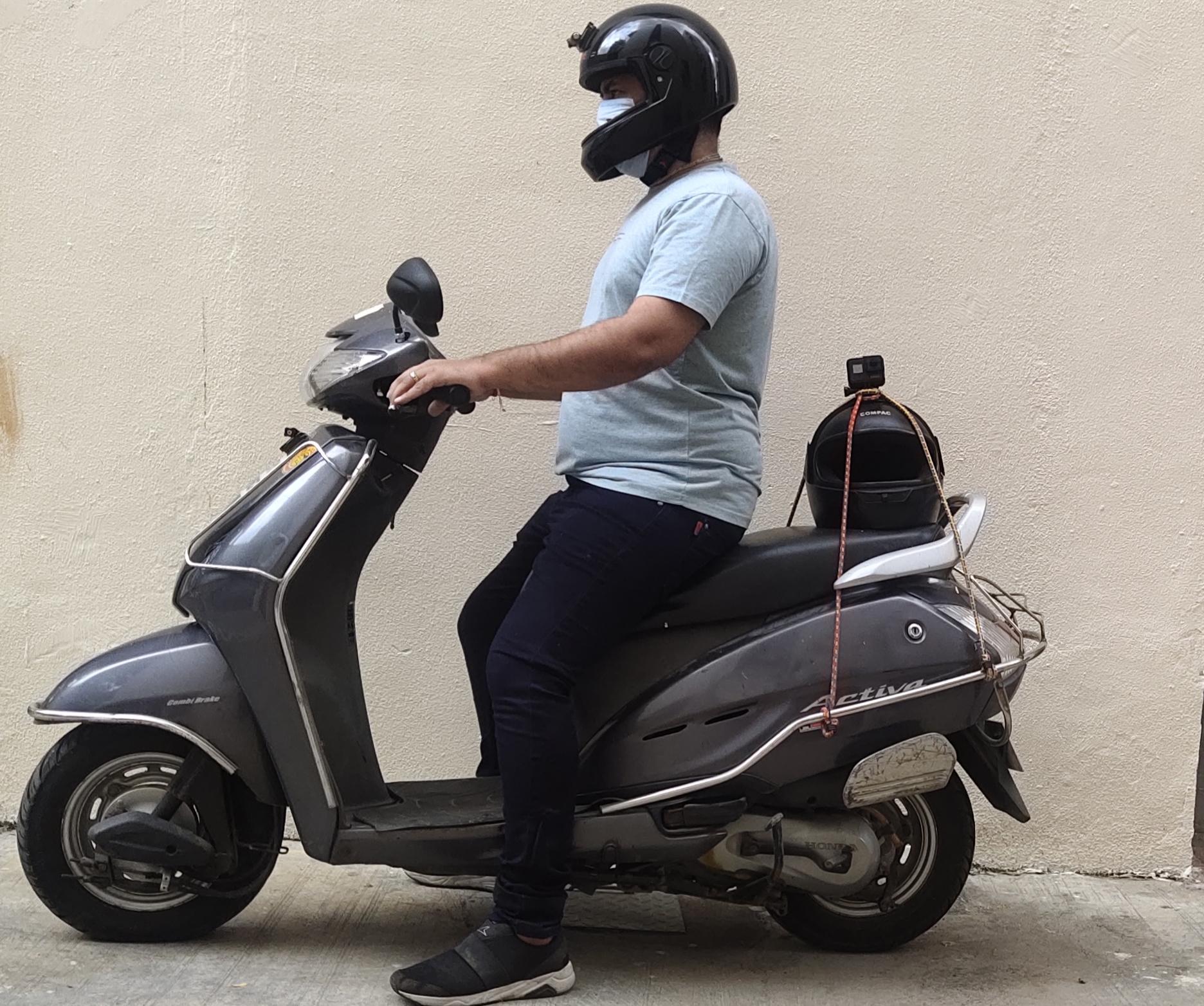} }}
    \qquad
    \subfloat[\centering]{{\includegraphics[width=0.38\textwidth]{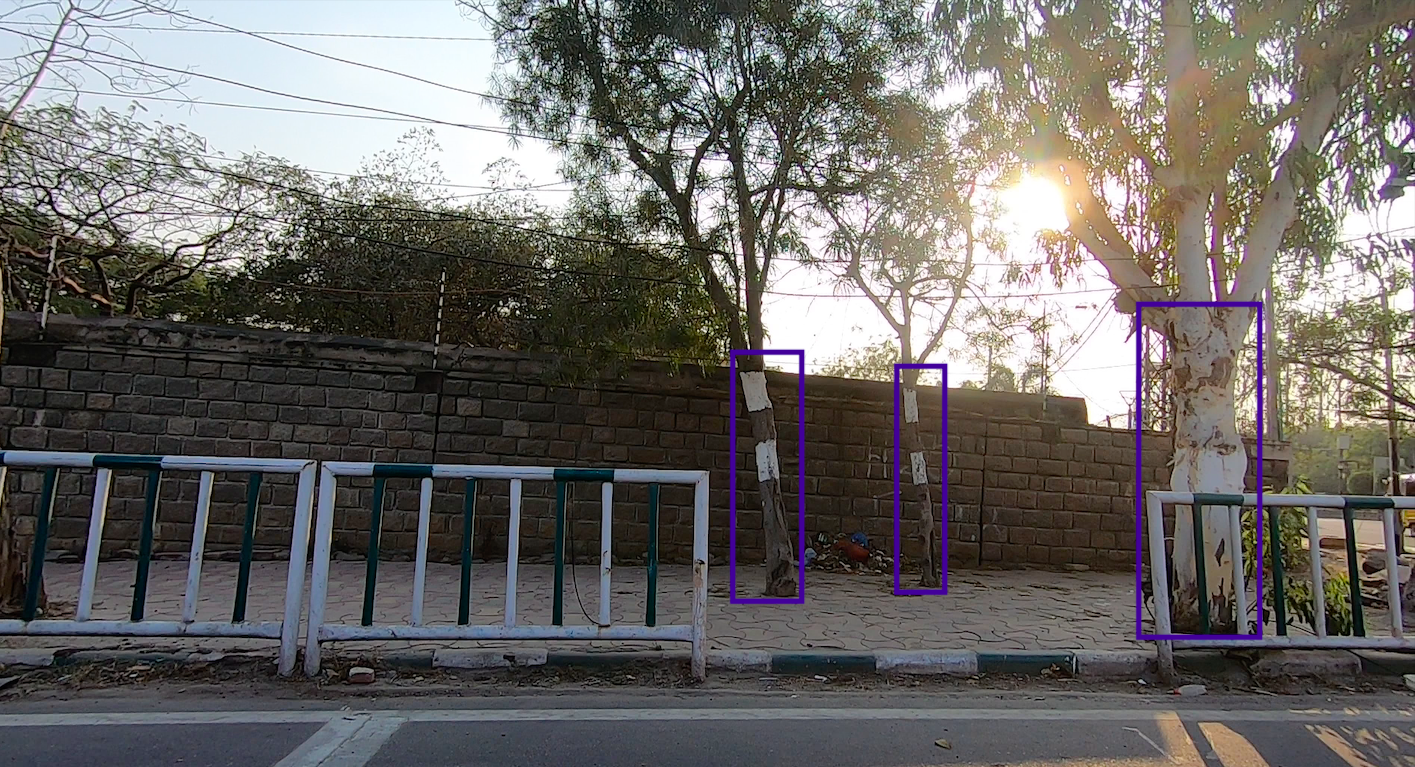} }}
    \vspace{-1em}
    \caption{(a) Our setup: we fix the camera over the top of a helmet mounted on the motorcycle's back seat, facing the sidewalk (when on the road). (b) Sample image, annotated with blue boxes, obtained from the camera.}
    \label{fig:setup&example}
    \vspace{-1em}
\end{figure*}
\section{Related Work}
We demonstrate related work in three categories. The first is the canopy cover estimation. The second is work related to tree detection, where the focus is only predicting boxes around trees. The third is tree assessment, where along with detecting trees, the trees are quantified, and results are showcased in a meaningful way to assess the trees.\\
\textbf{Canopy Cover Estimation}. Canopy cover estimation is a way to measure the greenery in cities. It is different from quantifying individual trees in a street. Existing methods that calculate canopy cover primarily rely on the original colour thresholding and clustering method to filter for possibly misidentified green specks~\cite{li, XuChen}. However, these methods tend to predict green objects with no vertical vegetation as green cover (false-positives). Moreover, they cannot predict non-green parts of vertical vegetation such as branches and yellow leaves as green cover (false-negatives). One of the most recent and popular works in canopy cover estimation that avoids the previously mentioned problem is by Cai et al.~\cite{cai}. They propose a semantic segmentation (PSPNet \cite{zhao}) and a direct end-to-end deep convolutional neural network to estimate Green View Index (GVI) to quantify urban canopy cover.\\
\textbf{Tree Detection}.
Earlier works on tree detection use Light Detection And Ranging (LiDAR)~\cite{chen,reitberger,lahivaara} or a combination of LiDAR and aerial imagery to detect trees~\cite{qin} and are mostly focused on delineating trees in forests. However, using LiDAR is costly, and obtaining such aerial images requires expensive flights or drone campaigns. One of the first works to use only camera based images for tree detection is by Yang et al.~\cite{yang}. In this work, the authors first classify aerial RGB images using a pixel-level classifier to a \{tree, non-tree\} label based on a set of visual features. In the second stage, they locate individual trees and provide an estimate of its crown size. However, this detection approach falls prey to locally indistinguishable objects such as grass and river. It remains unclear whether it will scale to entire cities with strongly varying tree shapes. Moreover, all the works mentioned hitherto have aerial views and not street views. Data in street view is important for assessing street trees because overhead imagery cannot represent the street-level and resident perspectives of street trees~\cite{yangJun}. Two relevant works that performed tree detection on street view images are by Itakura et al. and Xie et al.~\cite{itakura,xie}. Itakura et al.~\cite{itakura} performed automatic tree detection on street-view images from three-dimensional images reconstructed from 360-degree spherical camera using YOLOv2. Xie et al.~\cite{xie} are the closest to our work in terms of tree detection. The authors present a novel framework for tree detection, leveraging state-of-the-art deep learning based detection methods, along with two innovations in training loss definition and network module designation to handle the problem of inter-class and intra-class occlusion. They obtain their data from panoramic images captured from $22$ roads in the city of Nanjing in China using five fisheye lens. They take only two images on the left and right side from the panorama. However, in these works, the annotation process involves drawing the bounding box over a complete tree image, which carries complications (as we explain in Section~\ref{section:setup&dataset}). In contrast, we annotate only the trunks for detection. Section~\ref{section:setup&dataset} describes our method and its advantages.\\
\textbf{Street Tree Assessment}. Only detecting street trees would not provide any real-world application and thus, assessing street trees and showcasing the results in a meaningful way is important. One work that quantifies street trees is by Yao et al.~\cite{yao}. In this work, the authors aim to detect individual trees from satellite data and then perform tree counting based on density regression. They test four different Deep Neural Network (DNN) models on a tree counting dataset constructed with remote sensing images of 0.8m spatial resolution in distinct regions. However, satellite imagery is not conducive to assessing street trees, as mentioned previously. A different work by Branson et al.~\cite{branson} performs tree detection and cataloging using street view images. In this work, the authors propose an automated, image-based system to build up-to-date tree inventories at a large scale, using publicly available aerial images and panoramas at street-level. Two important things to note are --- First, they use multi-view detection. Secondly, the target outputs and training annotations in their object detection model (Faster RCNN) are geographic coordinates (latitude/longitude) rather than bounding boxes. They develop a technique to interchange between coordinates and bounding boxes. This work is closest to our work because they are also detecting individual trees from a street view. However, this work only pertains to the detection of street trees and building up the inventory. There is no tree counting and depiction of which areas have fewer trees and which have adequate. Moreover, they use publicly available street views that are not available for many places like India. Furthermore, they detect trees and build their inventory using static images. We take advantage of a vehicle driving and capturing the side view of the street and such a setup helps to count the trees efficiently.\\
To the best of our knowledge, ours is the first work that performs street tree quantification through street-view images and provides high-level visualizations of the number of trees in various city streets. Such quantification and visualizations can potentially help in urban planning and afforestation efforts. 
\section{Setup and Dataset}\label{section:setup&dataset}
\textbf{Setup}. Urban areas have trees in different types of regions, for example, roadside, urban forests or even parks. Many places in India have urban forests starting right next to a road. Thus, one of the challenges that we face is defining which trees would be assessed by our system. After a thorough analysis, we have developed a scheme that delineates the kind of trees counted and not counted by our system (Figure~\ref{fig:scheme}). As shown in this figure, we only count the trees that are on the roadside, and avoid counting plants, shrubs, trees in urban forests and trees behind walls or private properties. The data collection setup is an important part of our work. Our focus for the setup is to have low cost and good accuracy for detection and counting. The setup of the camera and an example image from the camera with annotation is shown in Figure~\ref{fig:setup&example}. Some key points of our setup are -
\begin{enumerate}
    \item  We mount a camera on a moving vehicle. The camera is facing towards the sidewalk\footnote{In countries with left-hand traffic (like India), the camera will be facing towards the left side, and in countries with right-hand traffic (like the USA), the camera will be facing towards the right side.}. Refer Figure~\ref{fig:scheme} for the field of view of the camera. Since the camera moves in a uniform direction and the object of interest (tree) is static, this setup is ideal for the counting algorithm we explain in Section~\ref{section:det&count}.
    \item Through a unique way of annotating trees, we tackle the problem of intra-class occlusion~\cite{xie}, where canopies of trees occlude each-other.
    \item By having the vehicle move on the lane closest to the sidewalk, we avoid inter-class occlusion (by other objects like vehicles, persons) considerably ~\cite{xie}.
    \item Public vehicles and surveillance vehicles can be used to ensure negligible additional costs. However, these vehicles may not always stay on the lane closest to the sidewalk, leading to a trade-off between cost and accuracy.
\end{enumerate}
Thus, our unique setup helps the tree detection model and the counting algorithm to work well. As we will see in Section~\ref{section:results}, we achieve accuracy close to human-level on a proposed metric.

\textbf{Dataset}. Existing object detection datasets with trees as one of the classes include OpenImages dataset~\cite{OpenImages}. The OpenImages dataset contains the bounding boxes annotations around the complete tree (including the trunk and the canopy). Annotating an entire tree creates a problem because the canopy is often not distinguishable, leading to ambiguity during annotation. However, the trunks are easily separable and thus provide a more straightforward and more consistent annotation task. An example of an instance where the canopies of multiple trees are not easily separable can be seen in Figure~\ref{fig:setup&example} (b). Here, it is difficult to mark the horizontal extents of the canopies of the three trees. 

Moreover, our work focuses on street tree quantification rather than just street tree detection. The extent of each tree is inconsequential for the problem of street tree counting. What is important is to count every tree and obtain high tree detection and counting accuracy, and for this, annotating just the trunk is favourable. Furthermore, annotating just the trunk also helps resolve the problem of intra-class occlusion because, unlike tree canopies, tree trunks are usually separable. Lastly, we also believe that detecting trees using only the trunks would help scale up to different species of trees better as the canopies of different species vary a lot, but their trunks are visually similar. However, we have not conducted any experiments on this as it requires additional annotation of tree classes by experts, and hence would be part of our future work. In this work, we focus on counting the trees, irrespective of their types.

Due to the above reasons, we develop a new dataset for street trees with the bounding boxes around the trunk. To collect the data, we first captured videos from a GoPro mounted on a motorcycle and driven on various roads of Hyderabad and Delhi. The video is captured at $30$ fps and at a resolution of $1080p$. To create the training dataset for the tree detection model, we then extract images from the videos at $1$ fps. There are $1295$ images with $1110$ images in the training set and $185$ images in the test set. To reduce the number of false positives (like poles, bus stand and pillars), we have carefully included $12\%$ of the entire training data as background images. We also have five test videos to evaluate counting results. Details about the dataset are given in Table~\ref{tab:dataset}.

\begin{table}
  \caption{Detailed information about the tree detection and counting dataset. The test images are to evaluate the detection results. The five test videos, covering approximately 25 km of roads, are to assess the counting results. The dataset is collected from streets in Hyderabad and Delhi. The resolution of all the images and videos is $1920\times1080$.}
  \vspace{-1em}
  \begin{tabular}{cccl}
    \toprule
  Group & \# Images/Videos & \# Trees \\
    \midrule
    Training Images & 1110 & 2265\\ 
    Test Images & 185 & 302\\ 
    Test Videos & 5 & 796 \\
  \bottomrule
\end{tabular}
\label{tab:dataset}\vspace{-1em}
\end{table}
Correct and consistent annotations are imperative for deep learning models. Since the tree trunks have inconsistent shapes, we develop certain annotation rules to have consistent bounding boxes. The top edge of the bounding box is made where the trunk starts branching out, or the trunk stops being visible, or where the leaves start appearing. For partially occluded trees, we only mark the visible part of the trunk. If other objects (humans, poles and so on ) bisects the trunk and occupies less than $15-20\%$ of the bounding box area, we include that in the bounding box. We exclude trees if less than $10-20\%$ of the object is visible, such that we cannot be sure if it is a tree. These images are annotated by a professional annotation lab using the Computer Vision Annotation Tool (CVAT) \cite{cvat}. An example of our tree annotation is shown in Figure~\ref{fig:setup&example} (b).
\begin{figure*}
 \makebox[\textwidth]{\includegraphics[width=0.92\textwidth]{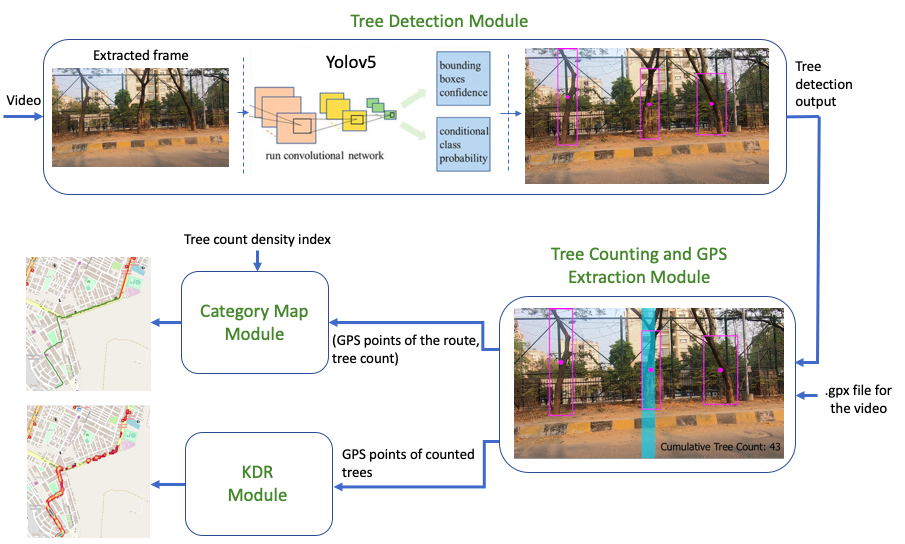}}\Description{Pipeline}
\vspace{-1em}
\caption{The pipeline of our system. The tree detection module detects trees in each frame of a video. In the tree counting and GPS extracting module, the detected trees are counted and their GPS locations stored. The blue box is the counting range (explained further in Section~\ref{section:det&count}). The cumulative tree count is updated in real-time in the video, as shown at the bottom right. Finally, the two visualization maps are shown at the bottom left.}
\label{fig:pipeline}\vspace{-1em}
\end{figure*}
\section{The System}\label{section:system}
The working of our system is as follows - we first pre-process the input video and then input it to a detection module. The captured videos could be long with a single video covering as much as 25 km and covering multiple streets. Since the end goal of our system is to provide meaningful interpretation in the form of maps of the tree density for individual streets, it is desirable to segment the videos into smaller distances. Thus, before the video is input into the detection module, we divide the input video into smaller segments. In the detection module, our object detection model outputs bounding boxes for detected trees. These bounding boxes along with a .gpx file are input to a tree counting and GPS extraction module. Our counting algorithm gives the cumulative count of trees at each frame and the final count at the end of the video. Moreover, whenever a tree is counted, the Global Positioning System (GPS) coordinates of the vehicle at that point are stored. Thus at the end of processing the video, we have fairly accurate GPS coordinates of the street trees. Now, to showcase the number of street trees in a meaningful way, we generate two different kinds of visualizations. First, the category map takes the following as input - tree count of the route, GPS coordinates of the route and our predefined scale (tree count density index, refer Section~\ref{section:categoryMap}) and outputs a map where the routes are coloured according to the index. Second, the density map takes the following as input - the GPS coordinates of the counted street trees and uses the Kernel Density Ranking (KDR) algorithm~\cite{kdr} to generate a map that showcases the density of street trees in a route. Note the detection and counting modules provide a separate output video with roadside tree detections and counts in mp4 format. The entire pipeline is showcased in Figure~\ref{fig:pipeline}.
\subsection{Detection and Counting}\label{section:det&count}
In the detection module, the model tries to predict bounding boxes for all instances of trees in each frame. It is important to note that we have only one class - tree. We train a YOLOv5  model \cite{yolov5} on the dataset described in the previous section. The task of tree detection is simplified due to the way we annotate trees. Since YOLOv5 is trained to detect only the trunks of trees and the trunks are usually separable, the detection model does not get confused in detecting individual trees where the canopies of those trees are not separable. Moreover, road scenes have other objects such as poles and brown coloured fences. that can be mistaken for a trunk and lead to false positive detections. To avoid this, we incorporate many such objects in our training dataset, and thus, the YOLOv5 model learns to distinguish trunks from other similar-looking objects.

There are various reasons for choosing YOLOv5 for tree detection. Firstly, YOLOv5 incorporates Cross Stage Partial Network (CSPNet) \cite{cspnet} into its backbone and in the neck. CSPNet helps to achieve a richer gradient combination while reducing the amount of computation, which ensures the inference speed and accuracy are high and reduces the model size. In the tree detection task, fast detection speed is imperative if the video capturing vehicle moves at high speeds. Moreover, to obtain accurate tree density maps, maintaining high accuracy is also essential. Furthermore, we also envision that this system will be deployed on resource-poor mobile devices. In such cases, compact models are essential. 
Secondly, the head of YOLOv5 generates $3$ different sizes ($18\times18$, $36\times36$, $72\times72$) of feature maps to achieve multi-scale~\cite{yolov3} prediction, enabling the model to handle small, medium, and big objects. Street trees usually have trunks of varying sizes in terms of width as well as height. Multi-scale detection ensures that the model can detect all such trees. 
YOLOv5 also auto-learns custom anchor boxes such that the anchors are particularly adapted to our tree dataset and this helps to improve the detection results. It also performs various augmentations such as mosaic augmentation during training which greatly helps to generalize. 
Moreover, we experiment with other models like - YOLOv4~\cite{yolov4} and Faster-RCNN~\cite{fasterRCNN} and we obtain the best results with YOLOv5 (refer Section~\ref{section:results}) while it also takes the least training time.

The predictions from the tree detection module are input to the tree counting and GPS extraction module. A tree is added to the count whenever the centre of the bounding box of a tree falls within a counting range. We define a counting range as a rectangular box with width as a fixed percentage of the image size and height equal to that of the image (refer filled blue rectangle at the bottom-right image in Figure~\ref{fig:pipeline}). The width of the counting range needs to be carefully decided. If the width is too low, system may miss counting a detected tree since its bounding box centre would never fall in the counting range. If the counting range is too broad, then the same tree would be counted multiple times which we term as double counting. It is easier to handle the problem of double count than missed counts. Thus, we ensure that the width of the counting range is large enough such that the center of the bounding box of a tree detection falls within it at least once in all the consecutive frames containing the detections of the tree (being counted). As a result, we will face the double count issue, which is solved using a simple technique that is explained next. We compare a newly detected bounding box in the counting range to certain previously counted bounding boxes. Suppose the new bounding box has a similar box size as any of the previously counted bounding boxes (within a predefined time period). In that case, we consider that it is the same object as one of the previously counted objects and has already been counted, so we ignore it. The bottom right image in Figure~\ref{fig:pipeline} illustrates the counting mechanism. The complete counting algorithm is explained in Algorithm~\ref{alg:1}. Furthermore, the counting and GPS extraction module takes a .gpx file corresponding to the input video. The GPS location of the entire route can be extracted from this file. In this module, we store the GPS locations of every tree at the frame in which it gets counted.
\begin{algorithm}[ht]
\SetAlgoLined
Let:

$box$ be a bounding box defined by $<$centre, height, width$>$\\
$bbs$ be list with each item as $box$, it contain all bounding boxes in a frame\\
$dict$ be a dictionary $<$box, frames\_counter$>$\\
$tree\_counter$ $\gets$ 0 \\
$iou\_th$ $\gets$ 0.5 \\
$next\_frames\_to\_consider$ $\gets$ 7 \\
\SetKwInOut{Input}{Input}\SetKwInOut{Output}{Output}
\Input{$\langle Video \rangle$} 
\For{$frame$ in $Video$}{
    \If{$dict$ is not empty}{
        \For{$key, value$ in $dict$}{
            $dict$[$key$]--\\
            \If{$dict$[$key$] == 0} {
                $dict$.pop($key$) 
            }
        }
    }
    \For{$bb$ in $bbs$} {
        $new\_obj$ $\gets$ True\\
        \If{$bb.centre$ lies in $counting\_range$} {
            \If{$dict$ is empty}{
                $dict[bb]$ $\gets$ $next\_frames\_to\_consider$\\
                $tree\_counter++$
            }
            \Else{
                \For {$key$, $value$ in $dict$} {
                    \If{IOU($bb$, $key$) $\ge$ $iou\_th$} {
                        $new\_obj$ $\gets$ False\\
                        break
                    }
                }
                \If{$new\_obj$ is True} {
                    $dict[bb]$ $\gets$ $next\_frames\_to\_consider$\\
                    $tree\_counter++$
                } 
            }
        }
    }
}
\vspace{0.5em}
\caption{Counting Algorithm. bb stands for bounding box, th for threshold and IOU for Intersection Over Union.}
\label{alg:1}
\end{algorithm}
\subsection{Count Density Index and Category Map} \label{section:categoryMap}
In this section, we explain how we develop the tree count density index and its interpretation. We also explain the first visualization map - category map (refer bottom-left image in Figure~\ref{fig:pipeline}). From the counting module we obtain a tuple of the GPS coordinates of the route and the count of trees in that route. We now create the category map using the tuple and a tree count density index that we develop and describe further. We create the index empirically by analysing the number of trees in thirty routes, each of length of 1 km. We create 5 bins according to the number of trees per km. Each bin has a corresponding interpretation of the street tree coverage ranging from very low tree coverage to very good tree coverage. The index is shown in Table~\ref{tab:treeCountIndex}. In case, we want to analyse routes of length different than 1 km, then the index is scaled linearly.

\begin{table}[ht]
  \caption{Tree Count Density Index}\vspace{-1em}
  \label{tab:treeCountIndex}
  \resizebox{\columnwidth}{!}{
  \begin{tabular}{cccl}
    \toprule
   Tree Count (per km)&Interpretation (category)&Colour on map\\
    \midrule
    < 20 & Very Low & Black\\
    20 to 30 & Low & Red \\
    30 to 40 & Moderate & Blue \\
    40 to 50 & Good & Green \\
    > 50 & Very Good & Dark green \\
  \bottomrule
\end{tabular}}\vspace{-1em}
\end{table}
\begin{figure}[t]
\includegraphics[width=\linewidth]{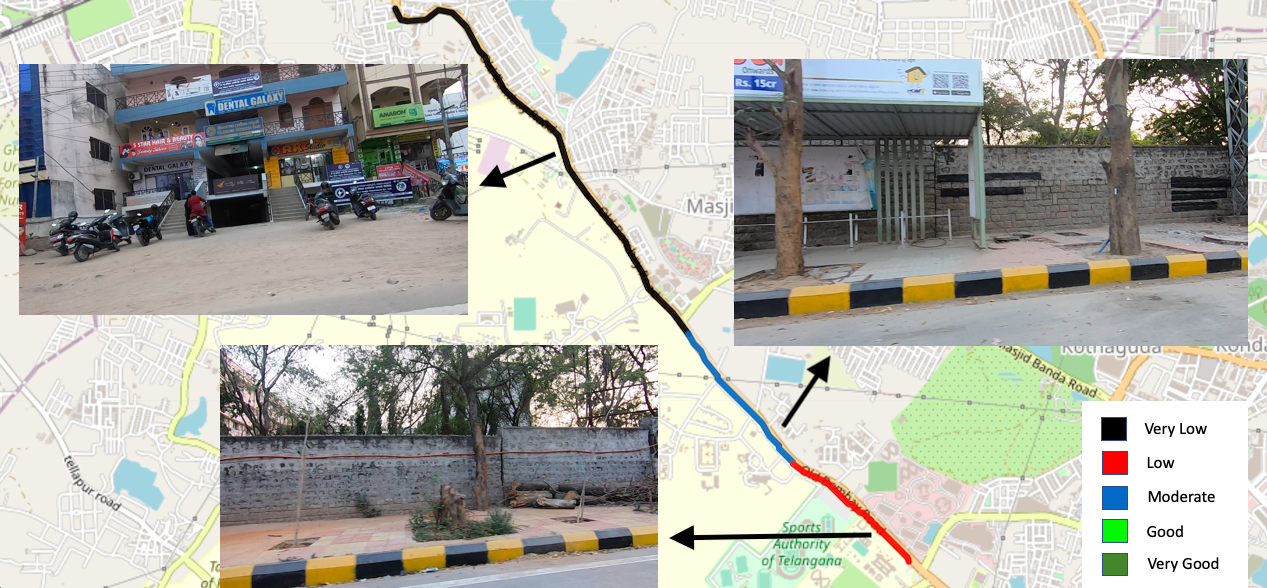}\Description{category map}
\vspace{-1em}
\caption{An example of a category map: the routes shown here are colour-coded to showcase the tree coverage in the corresponding streets. The images marked with the arrow indicate the scenes at particular instances on these routes. Note the image corresponding to the black route is tree-starved.}
\label{fig:categoryMap1}\vspace{-1em}
\end{figure}
Figure~\ref{fig:categoryMap1} shows how the category map looks like for sample routes in our training data. The advantage that the category map provides is that it helps to understand the high-level view of the number of trees in various streets of a city, town, or village. This in turn assists the policy-makers to quickly identify tree-starved roads and employ corrective measures to improve their green cover. One limitation of this map is when the tree distribution in a particular route is disparate. For instance, a route of $3$ km has a total tree count of greater than $180$, rendering this route a dark green. However, this route has a disparate distribution of trees such that, the first 1.5 km has 150 trees and the rest of the 1.5 km has just 30 trees. In this case, the category map would not showcase this disparity in the tree density. To solve this problem, we develop another map that we discuss in the next section, which showcases finer details about the tree densities in a route.

\subsection{Kernel Density Ranking and Density Map} \label{section:densityMap}

To interpret the results of a street tree detection and counting model, it is desirable to find out regions of high, low and moderate tree densities. Typically, Kernel Density Estimation (KDE) is used to estimate such densities. KDE allows us to estimate the probability density function from our finite dataset of detected street trees and corresponding GPS coordinates. However, KDE does not work well with GPS data~\cite{kdr}. Kernel Density Ranking (KDR)~\cite{kdr} is a better approach which is more conducive to GPS data. KDR is derived from KDE and thus we briefly explain the working of KDE first.

Let $n$ be the total number of trees detected and $\chi = \{X_1, X_2,..., X_n\}$ be the GPS positional data of each individual tree. The $i^{th}$ location is $X_i = (x_i,y_i)$ where $x_i$ and $y_i$ denote the latitude and longitude.

\textbf{Kernel Density Estimation}. This approach first partitions a wider area which consists of all the GPS coordinates in $\chi$ and partitions it into a raster grid. Each cell $x$, in this raster grid is assigned a value based on the distances from the center of the cell to the locations in $\chi$. The value of a grid cell $x$, is directly proportional to the density of trees in that cell. The estimate of the tree density at a grid point $x$ is given by
\begin{equation}
  \hat{p}(x) = \frac{1}{n{h}^2} \sum_{i=0}^{n}K \left(\frac{d_i(x)}{h}\right)
\end{equation}
Here $K()$ is a kernel function, $h$ is the bandwidth or smoothing parameter, and $d_i(x)$ is the distance between the grid point $x$ and the $i^{th}$ tree GPS coordinate $X_i = (x_i, y_i) \in \chi$. The most usual choice for $K()$ is a Gaussian function.  It is important to note that while calculating the tree density at a grid cell in the equation (1), we are considering the GPS points of all trees in $\chi$. However, this is not desirable as a tree GPS point that is at a large distance from a particular grid cell $x$, would not affect the density of that grid cell $x$. Thus, we choose a kernel function from Silverman et al.~\cite{silverman} that caters to the aforementioned problem and the KDE becomes
\begin{equation}
  \hat{p}(x) = \frac{3}{\pi{h}^2} \sum_{d_i(x)<h} \left(1- (\frac{d_i(x)}{h})^2\right)
\end{equation}
Thus, tree GPS locations in $\chi$ outside a circle with radius $h$ centered at $x$ are dropped in the evaluation of $\hat{p}(x)$. This ensures that when calculating the tree density at grid point $x$, the trees that are at a distance greater than $h$ do not contribute to it.

\noindent \textbf{Kernel Density Ranking}. KDR is defined as:
\begin{equation}
  \hat{\alpha}(x) = \frac{1}{n} \sum_{i}^{n} I\left(\hat{p}(X_i) \leq \hat{p}(x)\right)
\end{equation}
where $I(\omega)$ is the indicator function. The density ranking function $\hat{\alpha}(x)$ is the fraction of observations in $\chi = \{X_1, X_2, \ldots X_n\}$ whose KDE is lower than the KDE of the given point $x$. The density ranking function $\hat{\alpha}(x)$ is a probability-like quantity that takes values between 0 and 1. Density ranking has a straightforward interpretation. If a grid point $x$ has a value close to 1 that means that the tree density (KDE) at that grid point is greater than the tree density (KDE) of almost all the points in $\chi$. For instance, for a point $x$ with $\hat{p}(x) = 0.7$, the probability density (measured by the KDE $\hat{p}$) at point $x$ is higher than the probability density of $70\%$ of all GPS locations in $\chi$.

\begin{figure}[t]
\includegraphics[width=\linewidth]{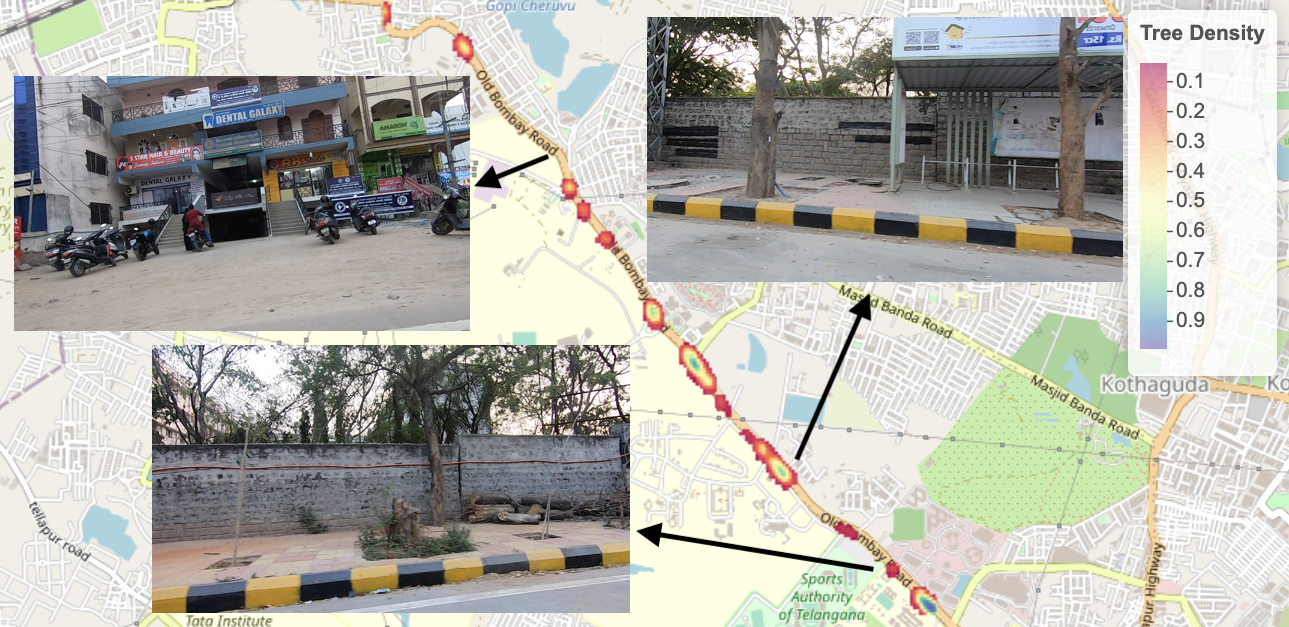}\Description{density map}
\vspace{-1em}
\caption{An example of a density map: the routes shown here are the same as in Figure~\ref{fig:categoryMap1}. The density of trees correspond to the colour of the corresponding route in Figure~\ref{fig:categoryMap1}. The images marked with the arrow show the real-world scene at certain instances on these routes.}
\label{fig:densityMap}
\vspace{-1em}
\end{figure}
Using the kernel density ranking algorithm, we create the density map for street trees. An example density map is shown in Figure~\ref{fig:densityMap}. The route shown in the density map in Figure~\ref{fig:densityMap} is the same as the route in the category map in Figure~\ref{fig:categoryMap1}. We can see that the colour of the route corresponds to the tree density in that route. Further, the density map provides finer details related to the tree density. For instance, for the black route in Figure~\ref{fig:categoryMap1}, the density map illustrates the exact locations of areas that are tree-starved and areas that have relatively more trees. The density map along with the category map, can help provide municipal or civic authorities with a high-level view of street tree densities of a city.

\begin{table}[t]
  \caption{Tree detection and counting results for various detectors. MSE is Mean Absolute Error, and TCDCA is Tree Count Density Classification Accuracy defined in Section~\ref{section:results}.}
  \label{tab:results}
  \vspace{-1em}
  \begin{tabular}{cccl}
    \toprule
   Model&mAP &  MAE & TCDCA\\
    \midrule
    Faster RCNN & 81.01\% & 6.12 & 74.19\%\\
    YOLOv4 & 82.50\% & 4.35 & 90.32\% \\
    YOLOv5s & 79.29\% & 7.22 & 67.74\%\\
    \textbf{YOLOv5l} & \textbf{83.74\%} & \textbf{3.09} & \textbf{96.77\%}\\
  \bottomrule
\end{tabular}
\vspace{-1em}
\label{table:results}
\end{table}
\begin{figure*}
    \centering
    \subfloat[\centering]{{\includegraphics[width=0.4775\textwidth]{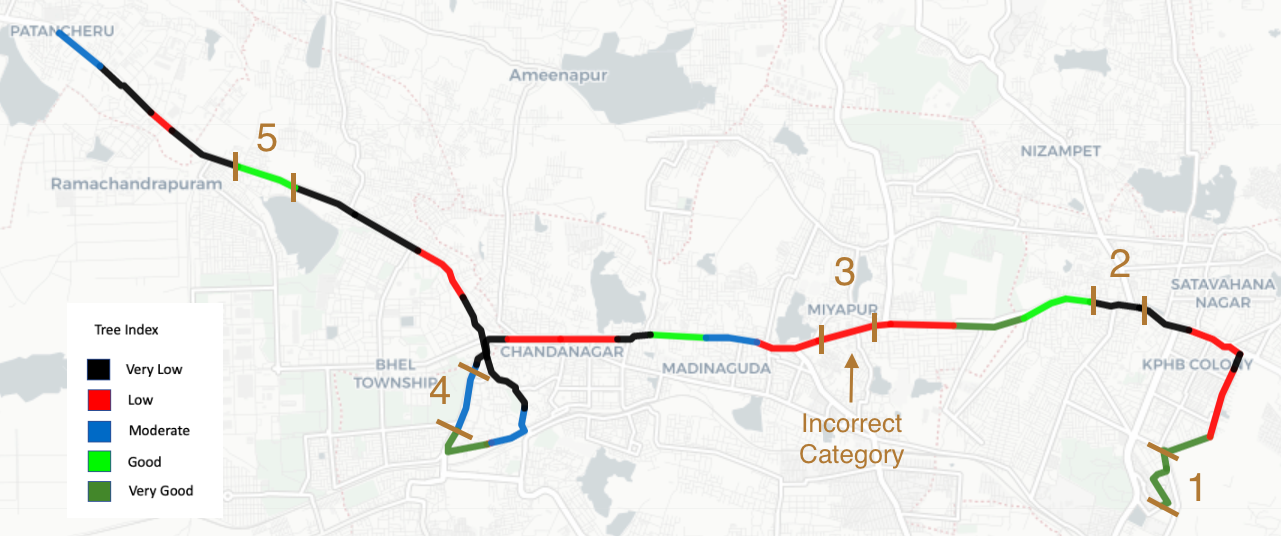} }}
    \qquad
    \subfloat[\centering]{{\includegraphics[width=0.4775\textwidth]{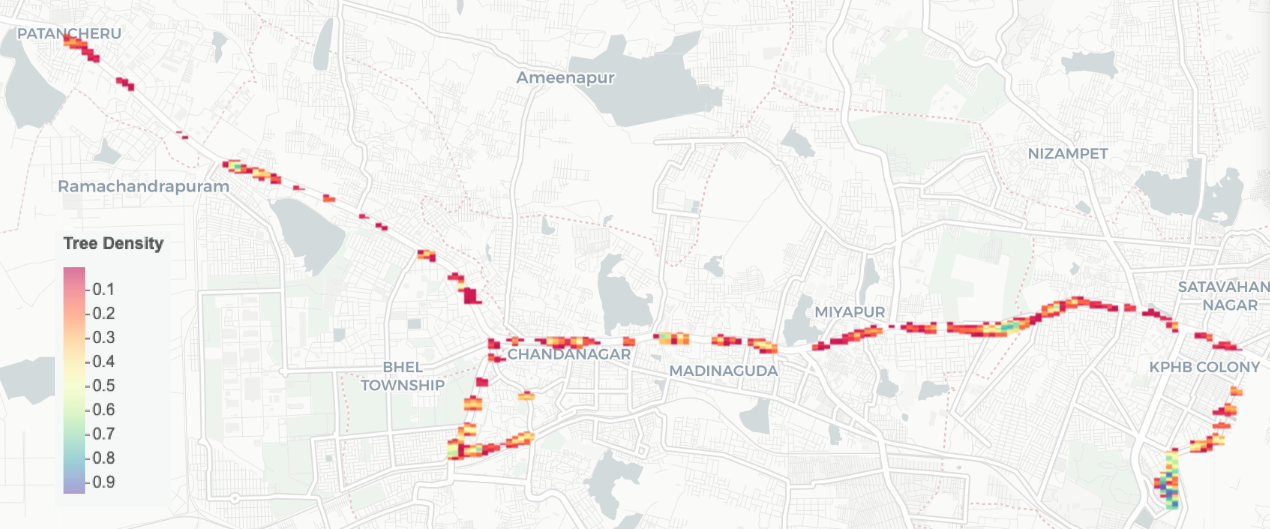} }}
    \vspace{-1em}
    \caption{(a) The category map based on the routes in our test data. The numbers on this map indicate the corresponding row for this route in Table~\ref{tab:countingResults} and the extent of these routes are marked by the brown lines. (b) The density map based on the routes in our test data, can be compared with the category map (a) to find that the colour of the route indeed corresponds to the tree density in that route.}
    \label{fig:mapResults}
    \vspace{-1em}
\end{figure*}
\section{Experiments and Results}\label{section:results}
We discuss the results of our detection and counting experiments in this section. Finally, we also present the obtained category and density maps on some of our test videos.\\
\textbf{Detection}
We train the YOLOv5l model for 100 epochs with a batch size of 4 on a GeForce GTX 1080 GPU. We use pre-trained weights on the COCO detection dataset. When training the YOLOv5l model with the default hyperparameters in the official YOLOv5~\cite{yolov5} codebase, we found that while the box loss on the validation set decreases, the objectness loss on the validation set reaches the lowest point and start increasing. To resolve this, we reduce the contribution of the objectness loss to the overall loss function by half. The mAP we obtain is $83.74\%$. We also try other models such as Faster RCNN, YOLOv4, and YOLOv5s, and none of these give a better mAP than YOLOv5l. The summary of the tree detection results for all the models is shown in Table~\ref{table:results}. Overall, we obtain a $2.73\%$ improvement of mAP over the baseline Faster RCNN model.\\
\textbf{Counting}. The test data for tree counting is different from that for detection as shown in Table \ref{tab:dataset}. The test set for tree counting consists of $5$ videos with a total of $25$ km being covered. These videos are first pre-processed and converted into smaller segments as discussed in Section \ref{section:system}. The ground truth of the tree counts for these routes is manually calculated by humans. We measure the counting performance of our models using two metrics: 
\begin{enumerate}
    \item Mean Absolute Error (MAE)
    \item Tree Count Density Classification Accuracy (TCDCA)
\end{enumerate}
Let $gt_{tree}^i$ and $p_{tree}^i$ be the ground truth and predicted tree count for the $i^{th}$ route. Let $N$ be the total number of routes. Then,
\begin{equation}
MAE = \frac{1}{N}\sum_{i=1}^{N}\parallel gt_{tree}^i - p_{tree}^i \parallel_1 
\end{equation}
As shown in Table~\ref{table:results}, we obtain a MAE of $3.1$ with YOLOv5l and achieve an improvement of $3.0$ in MAE over the baseline. This means that on average, the predicted count of trees varies from the true count of trees by an estimated value of $3.1$.

TCDCA measures how accurate is our system at predicting the correct category according to the tree count density index for individual streets. Let $R$ be the number of routes with correctly predicted category  and $N$ be the total number of routes. We define the TCDCA as $\frac{R}{N}$. We obtain a TCDCA of $96.77\%$ using our final YOLOV5l model, which is a remarkable improvement of $22.58\%$ over the baseline Faster RCNN model. This means that our system predicts the correct category for a route with a high accuracy. The category to which a street belongs to would help municipalities and civic authorities to identify which streets are tree-starved and which streets have ample trees. This in turn will help them to quickly deploy afforestation efforts. Our system can be used to identify such streets with a high reliability and with minimal costs. The MAE and TCDCA results for other models is shown in Table~\ref{tab:results}. Some interesting results are shown in Figure ~\ref{fig:interstingExample1}. We strongly recommend reviewers to see the supplementary video demo showing tree detection and counting results on test samples from our dataset.

\begin{figure}
\centering
    \subfloat[\centering]{{\includegraphics[trim={3cm 1.35cm 3cm 0cm},clip,width=0.215\textwidth]{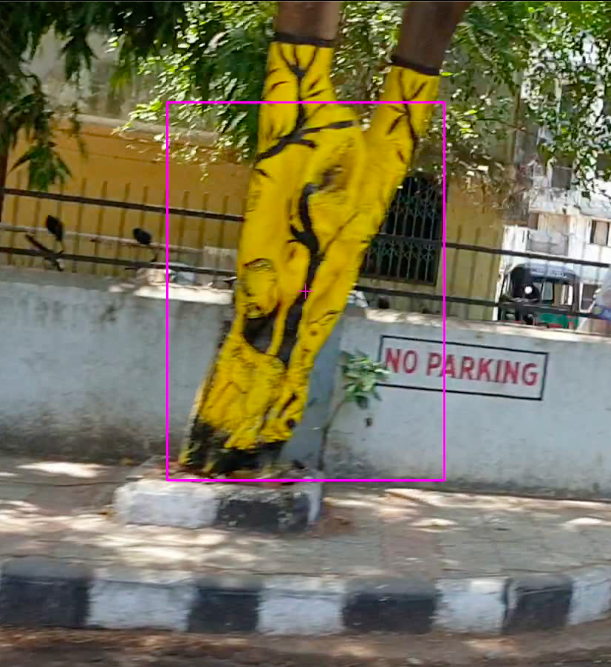} }}
    \qquad
    \subfloat[\centering]{{\includegraphics[trim={0cm 1cm 0cm 1.5cm},clip,width=0.215\textwidth]{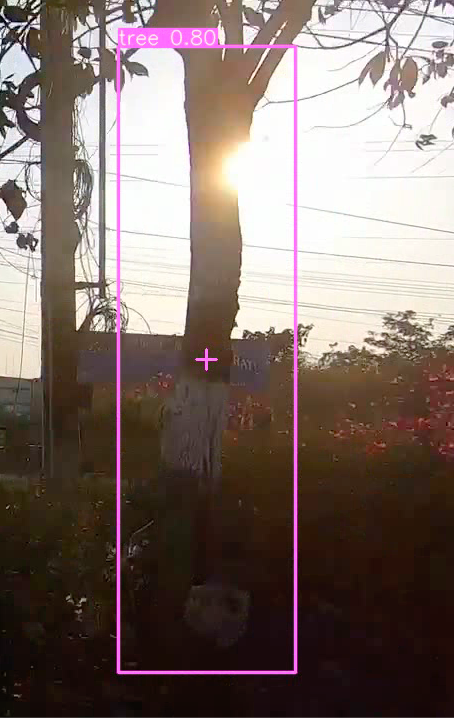} }}
\vspace{-1em}
\caption{Interesting examples: our model detects and counts (a) a tree whose trunk has been completely painted (such an example is not present in our training data) and (b) a tree with trunk image distorted due to glare from the sun.}
\label{fig:interstingExample1}
\end{figure}
\begin{table}[t]
  \caption{Few results for the tree counts and the corresponding category according to the tree count density index. The corresponding routes on a map is shown in Figure~\ref{fig:mapResults} (a). All of these routes are part of our test set.}\vspace{-1em}
  \resizebox{\columnwidth}{!}{
  \begin{tabular}{cccccl}
    \toprule
   Route&Distance (km)&GT Count&Pred Count&GT Category&Pred Category\\
    \midrule
    1 & 0.63 & 125 & 118 & Very High & Very High\\
    2 & 0.60 & 11 & 9 & Very Low & Very Low \\
    3 & 0.89 & 32 & 21 & Moderate & Low \\
    4 & 0.75 & 29 & 27 & Moderate & Moderate \\
    5 & 0.70 & 36 & 35 & High & High \\
  \bottomrule
\end{tabular}}
\vspace{-1em}
\label{tab:countingResults}
\end{table}
\textbf{Category Map and Density Map}. Figure~\ref{fig:mapResults} shows the resultant category and density maps for the routes that are part of our test set. We have indicated the actual count of trees for a few routes in Figure~\ref{fig:mapResults} (a) in Table~\ref{tab:countingResults}. As we can see in Figure~\ref{fig:mapResults} (a), only one route is incorrectly classified. We can also observe that the tree density in the density map corresponds to the colour of the route in the category map. For instance, route 2 indeed has less tree density than route 1 or 5. Thus, the category map helps to quickly, affordably and reliably identify tree-starved streets, and the density map provides finer details regarding tree-starved areas in that particular street. 

\section{Conclusion}
This paper presents a system for street tree detection, counting, and visualizing the tree distribution through two maps. Such a high-level visualization can help civic or municipal authorities identify tree-starved areas and employ afforestation efforts there. Specifically, we develop a unique setup and a street tree dataset with a novel annotation method, tweak and train a YOLOv5l model to detect street trees, and devise a new street tree counting algorithm. We achieve an mAP of $83.74\%$ and a tree count density classification accuracy of $96.77\%$. Further, to have meaningful results, we create two maps. The first is based on a tree count density index, an index that we develop to show the tree coverage on a route using colour codes on a map according to the tree count. The second is a map to showcase street tree densities.

Future direction of this work may include 1) incorporate new species of trees into our dataset, 2) upgrade the capability of our model to detect street trees that need protection, and 3) expand the testing of our system to other cities in India as well as outside India.\\

{\bf Acknowledgment:} We thank iHubData, the Technology Innovation Hub (TIH) at IIIT-Hyderabad for supporting this project.

\bibliographystyle{ACM-Reference-Format}
\bibliography{main}

\appendix

\end{document}